\documentclass{article}
\usepackage{ijcai24}
\usepackage{float}
\usepackage{times}
\usepackage{soul}
\usepackage{url}
\usepackage[hidelinks]{hyperref}
\usepackage[utf8]{inputenc}
\usepackage[small]{caption}
\usepackage{graphicx}
\usepackage{amsmath}
\usepackage{amsthm}
\usepackage{booktabs}
\usepackage{algorithm}
\usepackage{algorithmic}
\usepackage[switch]{lineno}
\usepackage{amsfonts,amssymb}
\usepackage{color}
\usepackage{bm}
\usepackage{tcolorbox}
\title{$\mathbf{C^3L}$: Content Correlated  Vision-Language Instruction Tuning Data Generation  via Contrastive Learning}

\author{
Ji Ma$^{1,2,3}$
\and
Wei Suo$^{1,2,3}$\and
Peng Wang$^{1,2,3}$\And
Yanning Zhang$^{1,2,3}$\\
\affiliations
$^1$School of Computer Science, Northwestern Polytechnical University, China\\
$^2$Ningbo Institute, Northwestern Polytechnical University, China\\
$^3$National Engineering Laboratory for Integrated Aero-Space-Ground-Ocean, China\\
\emails
\{maji, suowei1994\}@mail.nwpu.edu.cn,
\{peng.wang, ynzhang\}@nwpu.edu.cn
}

\begin{document}

\maketitle

\begin{abstract}
      Vision-Language Instruction Tuning (VLIT) is a critical training phase for Large Vision-Language Models (LVLMs).  
      With the improving capabilities of open-source LVLMs, researchers have increasingly turned to generate VLIT data by using open-source LVLMs and achieved significant progress. However, such data generation approaches are bottlenecked by the following challenges: 
      1) Since multi-modal models tend to be influenced by prior language knowledge, directly using LVLMs to generate VLIT data would inevitably lead to low content relevance between generated data and images.
      2) To improve the ability of the models to generate VLIT data, previous methods have incorporated an additional training phase to boost the generative capacity. This process hurts the generalization of the models to unseen inputs (\emph{i.e.,} ``exposure bias'' problem).
      In this paper, we propose a new \textbf{C}ontent \textbf{C}orrelated VLIT data generation  via \textbf{C}ontrastive \textbf{L}earning ($C^3L$). 
      Specifically, we design a new content relevance module which enhances the content relevance between VLIT data and images by computing \textbf{I}mage \textbf{I}nstruction \textbf{C}orrespondence \textbf{S}cores $S(I^2C)$.
      Moreover, a contrastive learning module is introduced to further boost the 
      VLIT data generation capability of the LVLMs.
      A large number of automatic measures  on four benchmarks show the effectiveness of our method.\footnote{https://github.com/Fake10086/C3L}

\end{abstract}

\section{Introduction}
\begin{figure}[h]

\centering

\includegraphics[width = 8.5cm ]{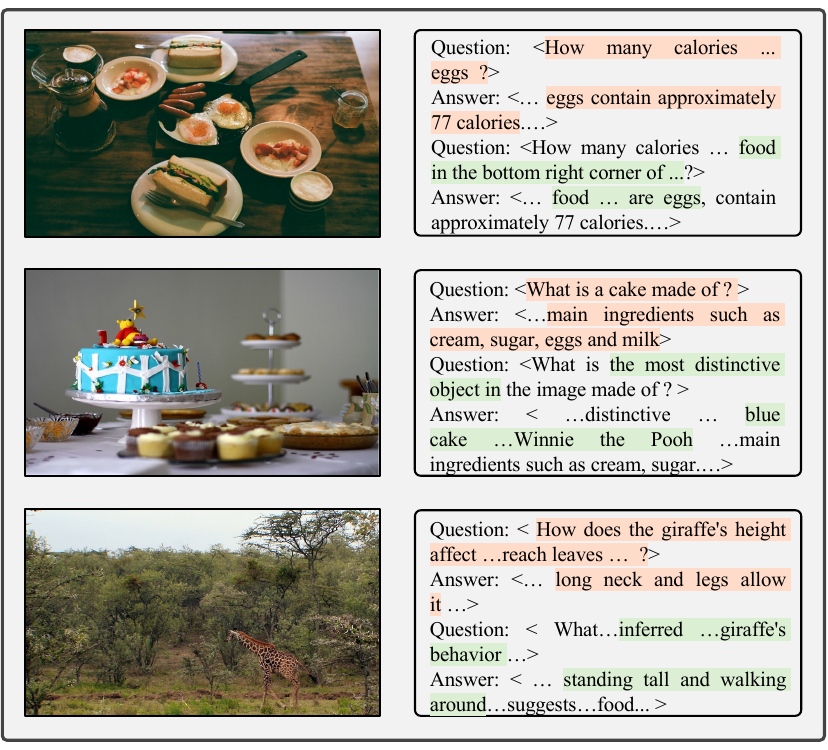}
\caption{The illustration of the prior language knowledge problem when directly using current LVLMs to generate VLIT data. Existing models tend to generate data that  exhibits low content relevance with the corresponding images (denoted in red). 
 Our method effectively enhances the content relevance between VLIT data and images (denoted in green).
}
\label{fig:fig_1}
\end{figure}
In recent years, significant advancements have been made in the field of natural language processing, with the emergence of Large 
Language Models (LLMs) revolutionizing the landscape of this field~\cite{openai,palm,llama}.
Leveraging the powerful reasoning capabilities of LLMs, researchers~\cite{blip2,minigpt4,llava} have proposed integrating visual encoders with LLMs to construct multi-modal perception and reasoning systems. This integration empowers LLMs with the ability to perceive and process visual information,
 leading to the substantial strides of Large Vision-Language Models (LVLMs).

In practice, most of the existing LVLMs adopt a two-stage training paradigm~\cite{llava,minigpt4,llava-v1-5,blip2,instructblip}. In the first stage, a substantial amount of image-text pairs are used to pre-train LVLMs, typically employing the image-text contrastive approach~\cite{llava,llava-v1-5,minigpt4}. The objective of this pre-training phase is to develop the fundamental cross-modal alignment capabilities of LVLMs.

In the second stage, a shift is made from traditional task-specific fine-tuning methods to a more general approach~\cite{vlit-survey}. Instead of relying on task-specific data for individual downstream tasks~\cite{bert,t5}, models are now fine-tuned using high-quality Vision-Language Instruction Tuning (VLIT) data. This approach aims to develop  general abilities of the models to understand and follow various types of instructions while generating helpful, factual, and harmless responses.
The primary approach~\cite{llava,instruct-gpt4v} to generating VLIT data is through using GPT-4(V)~\cite{gpt} to reduce  manual annotations. 
With the improving capabilities of open-source LVLMs, researchers are increasingly applying these models to generate VLIT data, enabling them to overcome the limited accessibility of GPT-4(V)~\cite{vigc,qwen-vl,vlit-survey}.
\begin{figure*}[h]

\centering

\includegraphics[width = 17.5cm]{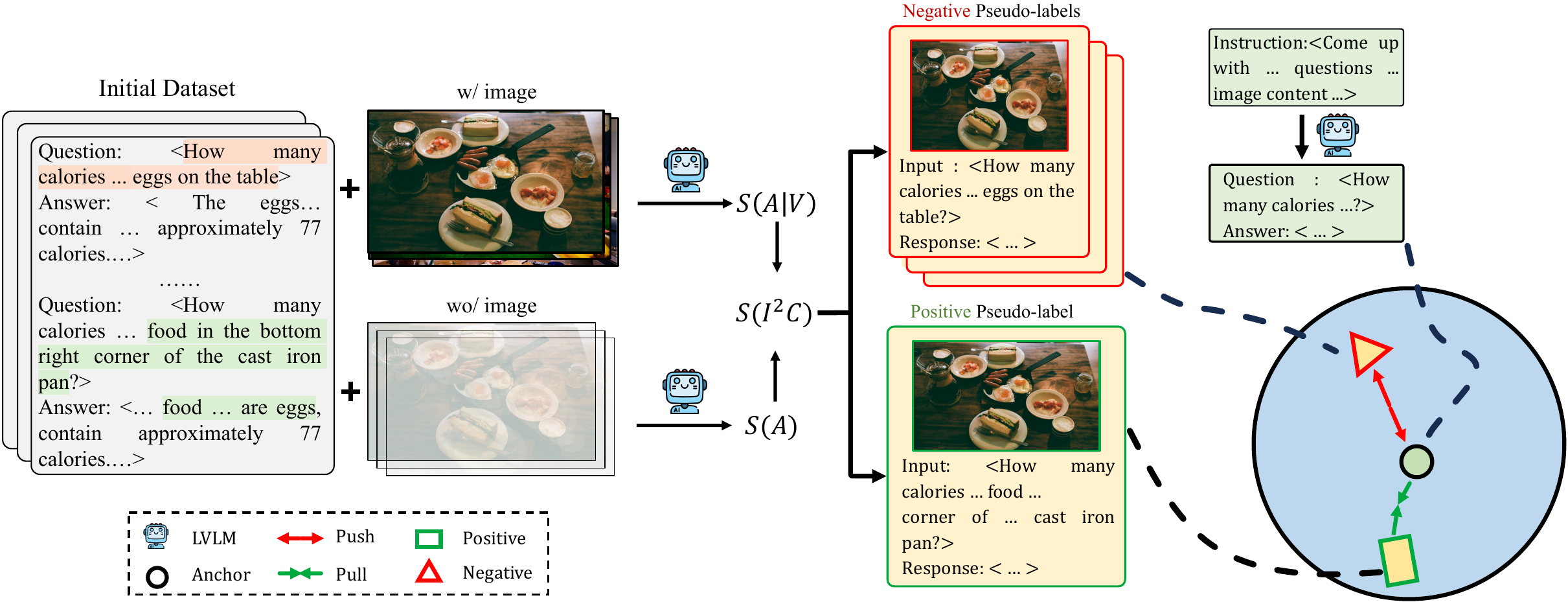}
\caption{Overview of our Content Correlated VLIT data generation via Contrastive Learning (\textit{C$^3$L}). Given the initial dataset and corresponding images, we first use Content Relevance module to obtain the $I^2C$ scores based on whether or not the image is provided. Then, positive-negative pseudo-labels are selected based on $I^2C$ scores. Further,  our Contrastive Learning module maximize the similarity between the anchor and positive pseudo-label while minimizing the similarity between the anchor and negative pseudo-labels.}
\label{fig: fig_2}
\end{figure*}

Although significant progress has been made, this paradigm still faces the following challenges: 
1) A longstanding issue with multi-modal models is to overly depend on prior language knowledge~\cite{make-v-matter},  leading them to ignore visual content. 
Therefore, when generating VLIT data using LVLMs, the models often fail to effectively focus on the visual information and instead heavily rely on prior language knowledge. As shown in Fig. \ref{fig:fig_1}, the resulting VLIT data (denoted in red) demonstrates limited correlation with the corresponding images.
2) Current LVLMs are primarily trained with an emphasis on developing strong reasoning capabilities rather than their abilities as data generators. 
In fact, a straightforward approach~\cite{vigc} to address this issue is by introducing an additional training phase where the models are trained with high-quality VLIT data as ground truth labels.
As evidenced in~\cite{exposure-bias,contrastive}, models that are only exposed to correct   samples suffer from the “exposure bias” problem, which will restrict generalization abilities of the models when encountering 
 unseen samples. 
Therefore, this training paradigm hinders the effectiveness of turning the models into  data generators.

To address the aforementioned challenges, we propose a new \textbf{C}ontent \textbf{C}orrelated VLIT 
data generation via \textbf{C}ontrastive \textbf{L}earning, which is called $C^3L$ for short. 
In $C^3L$, we first apply the LVLM to generate a set of initial VLIT data.
Then, to enhance the content relevance between VLIT data  and images, we propose a novel content relevance module where \textbf{I}mage \textbf{I}nstruction \textbf{C}orrespondence ($I^2C$) scores are computed based on whether or not the images are provided. 
On the other hand, due to the training paradigm that only utilizes high-quality VLIT data as ground truth labels lacks exposure to  low-quality samples, we divide the initial VLIT data into two categories based on the $I^2C$ scores. The data samples with high $I^2C$ scores are considered as positive pseudo-labels, and vice versa. By employing contrastive learning framework~\cite{contrastive} with positive-negative pseudo-labels, we can effectively alleviate the “exposure bias” problem and enhance the  capability of the model as a data generator.
Benefiting from the above methods, $C^3L$ effectively improves the content relevance of the model-generated VLIT data with images and enhances the  capability of the model as a data generator. 

According to automatic evaluations, fine-tuned LVLMs using data generated by  $C^3L$ achieve comparable  performance to the state-of-the-art models on four recent multi-modal benchmarks. More importantly,  only 5k VLIT data samples generated by our $C^3L$ are used to fine-tune these models, significantly reduces the computational cost. In summary, we make the following contributions:

1) We propose a new  content relevance module that can model the content correlation between VLIT data and images. This module effectively improves the content relevance and the utilization of visual information.

2) We develop an advanced contrastive learning module for VLIT data generation, which applies generated samples as pseudo-labels to boost the data generation capacity of the LVLMs further. To the best of our knowledge, we are the first to explore contrastive learning on the VLIT data generation.

3) The LVLMs fine-tuned using data generated by our  $C^3L$ achieves comparable or even better performance on four multi-modal benchmarks (\textit{i.e.,} SEED~\cite{seed-bench}, LLaVA$\rm^W$~\cite{llava-v1-5}, MMB~\cite{mmbench} and POPE~\cite{pope}). Meanwhile, automatic measures and ablation studies all show the effectiveness of our method.
\section{Related Works}

\subsection{Large Vision-Language Models}
Most current LVLMs typically comprise three main components~\cite{lvlm-survey-1,lvlm-survey-2}: visual encoder, LLM and projection layer to connect visual encoder and LLM. 
In order to  connect the visual encoder and the LLM, a connectivity structure is employed~\cite{instructblip,llava,minigpt4}.
Typically, the majority of LVLMs  adopt a two-stage training paradigm, encompassing Vision-Language Pre-training (VLP) and VLIT. 
In the first stage, large scale image-text pairs are employed~\cite{blip2,vision-language-pretrain} to establish  fundamental vision-language alignment of LVLMs. This process involves training models to comprehend the visual information in the images and generate  captions that accurately depict the visual content. 
In the second stage, a significant corpus of instruction data generated by models~\cite{qwen-vl,instruct-gpt4v} is utilized to enhance the capacity of LVLMs in comprehensively understanding vision-language instructions and generating appropriate responses. 
 Existing LVLMs tend to excessively rely on prior language knowledge~\cite{hallusionbench}. Therefore, when employing them to generate VLIT data, the generated data exhibits limited content relevance with images. In order to alleviate this issue, we design a novel content relevance module to improve the content correlation between model-generated VLIT data and corresponding images.

\subsection{VLIT Data Generation}
In order to enhance the capabilities of LVLMs to understand and follow instructions, the VLIT phase based on VLIT data has been introduced~\cite{vlit-survey,lvlm-survey-1}. 
To reduce the cost of human annotation, the generation of VLIT data primarily relies on automatic model generation~\cite{vlit-survey}. 
This is achieved by utilizing closed-source, powerful large models such as GPT-4(V)~\cite{llava,instruct-gpt4v}. 
Specifically, ~\cite{llava}  collects VLIT data based on the existing image-pair datasets. By prompting text-only GPT-4~\cite{gpt}, high-quality question-answer pairs are obtained and used as VLIT data (\textit{i.e.,} LLaVA-158k). Moreover, as generating VLIT data solely by prompting text-only models like GPT-4 would inevitably result in the loss of  visual information, previous work~\cite{instruct-gpt4v} proposes to leverage GPT-4V~\cite{gpt} to generate VLIT data with the entire visual context.
Recently, as open-source LVLMs continue to evolve, researchers~\cite{qwen-vl,vigc} have turned to generate  VLIT data  by LVLMs. ~\cite{qwen-vl} employs self-instruction to acquire VLIT data, which can be used to improve the image content comprehension ability of the LVLMs.  
These approaches contribute to further reducing the cost associated with assessing GPT-4(V).
However, as current open-source LVLMs are primarily trained with an emphasis on developing strong reasoning abilities, directly applying the models  to generate
VLIT data may result in undesirable results.
Different from the above methods, we introduce a new contrastive learning module to boost the data generation capacity of the LVLMs.

\subsection{Contrastive Learning}
Contrastive learning proposed in~\cite{cl-2006} has been demonstrated to be an effective method for visual feature extraction~\cite{cl-2021}. ~\cite{cl-2020} shows that contrastive learning benefits from larger batch sizes and can boost the performance of self-supervised learning in computer vision tasks.
Some studies have also applied contrastive learning to multi-modal sequence modeling tasks, such as image captioning~\cite{cl-4-image-caption} and visual question answering~\cite{cl-4-vqa-1,Multi-level-contra,s3c}.  
However, these studies primarily focus on utilizing contrastive learning methods to improve the multi-modal reasoning abilities and robustness of models. In contrast, our contrastive learning module is applied to enhance the  capability of the models in generating VLIT data that exhibits higher content relevance with images. By employing positive-negative pseudo-labels with the contrastive learning framework, our contrastive learning module can alleviate the “exposure bias” problem~\cite{exposure-bias} which hurts the generalization of the models and thereby transform the model into a more capable data generator.

\section{Method}
The aim of this paper is to  generate  VLIT data that exhibits high content relevance with corresponding images and enhance the capability of the model as a data generator. 
We achieve this by addressing two issues that prior works have ignored:
1) Due to the tendency of multi-modal models to overly rely on language prior knowledge~\cite{make-v-matter}, the relevance between the 
 model-generated data  and  images is limited.
2) Considering the low capacity of  current LVLMs in generating data, VIGC~\cite{vigc} introduces an additional training phase to transform the model into a data generator using  high-quality VLIT data as ground truth labels. This  training paradigm has been demonstrated to result in the “exposure bias” problem~\cite{exposure-bias,contrastive}.
In this section, we introduce our \textbf{C}ontent \textbf{C}orrelated VLIT data generation  via \textbf{C}ontrastive \textbf{L}earning ($C^3L$). As shown in Fig. \ref{fig: fig_2}, the $C^3L$ comprises 
an augmented pipeline with content relevance module and contrastive learning module.
Next, we would
introduce  our method in detail.

\subsection{Conventional VLIT Data Generation Pipeline}
\paragraph{VLIT data.}  
Given the image $I_i$, the VLIT data  is structured as  question-answer pairs $[Q_j,A_j]$, where $j \in \{1,2,...,N_j\}$  
denotes the $j$-th question-answer pair associated with the  image $I_i$. 
\paragraph{Conventional  pipeline.} 
As in~\cite{vigc}, the LVLM that used for generating VLIT data  is first trained  with high-quality VLIT data as ground truth labels. 
Next, given a set of instructions and images, the enhanced LVLM is utilized to generate VLIT data which is  in the form of question-answer pairs 
$[Q_j,A_j]$. The instructions here are used to guide the enhanced LVLM in generating VLIT data. Finally, the data can be used to fine-tune LVLMs in order to enhance the general capabilities of the models.

\paragraph{Limitations.} 1) The process overlooks the impact of language priors~\cite{make-v-matter} embedded in LLM counterparts of LVLMs, resulting in limited correlation between the generated VLIT data and the corresponding images. 
2) The training phase only
uses high-quality VLIT data as ground truth labels. As evidenced in~\cite{contrastive}, this training paradigm suffers from the “exposure bias” problem~\cite{exposure-bias}  and hinders the effectiveness of transforming the model into a data generator.

\subsection{Content Relevance Module}
\label{sec:sec_3_2}
To solve the first issue,  we propose a novel  content relevance  module which computes the \textbf{I}mage \textbf{I}nstruction \textbf{C}orrespondence \textbf{S}cores $S(I^2C)$ based on whether the image is present or not. 

Given a set of  instructions and  images, we first apply a widely used LVLM such as LLaVA~\cite{llava}  to generate the initial VLIT data. 
Due to the influence of language prior knowledge, the initial dataset  contains data samples that exhibit low relevance to the image content, as shown in Fig. \ref{fig:fig_1}. 
Next, for a given data sample $[Q_j,A_j]$ in our initial VLIT dataset, 
we set $Q_j$ and the corresponding image as inputs to the model. After obtaining the outputs of the model, we compute the probability value $p^v_t$ of the outputs for each answer token in $A_j$, where $t$ denotes the $t$-th token. Then, these probability values are concatenated to obtain the Visual Answer Scores $S(A|V)$.The $S(A|V)$ scores measure the  responses of the LVLM to a specific VLIT data given the image. 
We can also compute the probability values $p_t^d$ and obtain the Direct Answer Scores  $S(A)$ in the absence of the image.
The  $S(A)$ scores quantify the  responses of the model to a given VLIT data in the absence of an image.
The final  scores $S(I^2C)$ is computed  by using KL-divergence:
\begin{equation}
\begin{aligned}
    S(I^2C) =& D_{KL}(S(A|V)||S(A)) \\
    =& \sum_{t=1}^n p^v_t \cdot \log\frac{p^v_t}{p^d_t}.
\end{aligned}
\end{equation}
The  $S(I^2C)$ scores measure the differences in  responses of the model when provided with and without an image.
Based on the properties of KL-divergence~\cite{kl-div},  a low $S(I^2C)$ means that the divergence between  $S(A|V)$ and $S(A)$ is  small, indicating that the content relevance between the data sample and the image is low.

After obtaining the $S(I^2C)$ scores for all initial VLIT data samples, we  train the model  by using 10 $\%$ data samples with higher $S(I^2C)$ as ground truth labels, the standard cross-entropy loss $L_r$ is used during training. 

\subsection{       Contrastive Learning Module}
Due to the presence of the “exposure bias” problem~\cite{exposure-bias} when training models solely with high-quality VLIT data as ground truth labels, we introduce an additional contrastive learning module in which the model also learns experience from low-quality data samples in the initial VLIT dataset.
Following contrastive learning framework~\cite{contrastive}, we maximize the similarity between the  anchor and positive pseudo-labels while minimizing the similarity between the anchor and negative pseudo-labels as follows. 

In particular, given an image $I_i$, 
  the data sample $[Q_j, A_j]$ with the highest $S(I^2C)$ is selected and employed as the positive pseudo-label.  We  concatenate $[Q_j,A_j]$ and
embed the sequence $s$ using text encoder in the LVLM, denoted as $\mathbf{e_s} \in \mathbb{R}^{d\times l}$ where $l$ represents the length of sequence.
Affine transformation $\xi$ with the ReLU and AvgPool is  used to project $\mathbf{e_s}$ onto the latent space
$\mathbf{h_s} \in \mathbb{R}^{d}$. This process is denoted as:
\begin{equation}
\begin{aligned}
    &\mathbf{e_s} =  {\rm Embedding}(Q_j, A_j), \\
    &\mathbf{h_s} = \xi(\mathbf{e_s}),
\end{aligned}
\label{eq:eq_4}
\end{equation}
where $\xi(\cdot) = {\rm AvgPool}({\rm ReLU}(\cdot))$. 

Similarly,  the remaining data samples $\mathbf{\hat{s} \in S}$ corresponding to the image $I_i$ are selected and employed as the negative pseudo-labels. 
The  representations $\mathbf{h_{\hat{s}}}$ are computed  using eq.\ref{eq:eq_4}. Then, we input an instruction  along with the image $I_i$ into the model and obtain the anchor data ${y}$. 
Finally, we compute the  contrastive loss $L_c$ as follows:
\begin{align}
    L_c = -\log\frac{exp(sim(\mathbf{h_s},\mathbf{h_y})/\tau)}{\sum\nolimits_{\hat{s}\in S}exp(sim(\mathbf{h_{\hat{s}}},\mathbf{h_y})/\tau)},
\end{align}
where $sim(\cdot,\cdot)$ is a cosine similarity function.

\subsection{Augmented VLIT Data Generation Pipeline}
By using our content relevance module and contrastive learning module, we can construct an augmented VLIT data generation pipeline.

\paragraph{Initial data generation.} Given a set of instructions and images, we first employ a widely used LVLM such as LLaVA~\cite{llava} to generate  a set of initial VLIT data, wherein the data is in the form of question-answer pairs. Inspired by~\cite{changpinyo2022image_caption_for_vqa}, we use a simple two-step generation (\textit{i.e.,} first generate captions for the images and then use the LLM-counterparts of LVLMs to generate data based on captions) to further fill the gap between the modalities. 
Following~\cite{vigc}, the instructions are primarily categorized into three main classes: 1) conversation 2) detailed description 3) complex reasoning. The instructions are used to guide the model in generating VLIT data (\textit{e.g.,} \textit{ Generate five in-depth reasoning questions and then answer them based on the image.} or \textit{Generate five questions in order to describe the image and then answer them.}).

\paragraph{Content  relevance  module.}
Due to the language priors embedded in LLM counterparts of LVLMs, the initial VLIT dataset contains data samples that exhibit low content relevance between the VLIT data and the corresponding images. 
By using our content  relevance  module, we can compute the $S(I^2C)$ for all VLIT data samples based on whether or not the image is provided and train the model by using 10 $\%$ data samples with higher $S(I^2C)$ as ground truth labels. The standard cross-entropy loss $L_r$ is used during this training phase. 

\begin{table*}
    \centering

\resizebox{\linewidth}{!}{\begin{tabular}{{l}|cc|cccc}
\toprule
 Method  & Data & Base LLM & SEED  & LLaVA$\rm^W$ & MMB & POPE \\
\midrule
MiniGPT-4~\cite{minigpt4} & 3.5k & Vicuna-7b & 43.3 & - & 24.3 & 74.1  \\ 
VisualGLM~\cite{visualglm} & - & ChatGLM-6b & 47.0 & - & 37.6 & -\\ 
LLaVA~\cite{llava} & 158k & Vicuna-7b & 49.4 & - & 49.9 & 67.8  \\ 
BLIP2~\cite{blip2} & - & Vicuna-13b & 46.4 & 38.1 & - & \textbf{85.3}  \\ 
InstructBlip~\cite{instructblip} & 1.2M & Vicuna-7b & 53.4 & 60.9 & 36.0 & 83.7 \\ 
InstructBlip~\cite{instructblip} & 1.2M & Vicuna-13b & - & 58.2 & - & 78.9 \\ 
IDEFICS-80B~\cite{IDEFICS} & 1M & Llama-65b & 52.0 & - & \textbf{54.6} & - \\
\midrule
MiniGPT-4 w/ $C^3L$$^*$ & 5k & Llama-7b & 44.1 &  59.7 & 42.6 & 70.3 \\ 
MiniGPT-4 w/ $C^3L$ & 5k & Llama-7b & 40.9 & \textbf{63.6}  & 37.0 & 70.4 \\ 
LLaVA w/ $C^3L$$^*$ & 5k & Vicuna-7b & 52.7 & 62.2 & 47.6 & 75.7  \\ 
LLaVA w/ $C^3L$ & 5k & Vicuna-7b & \textbf{56.1} & 61.7 & 
51.9  & 76.0  \\
\bottomrule
\end{tabular}}
    \caption{Comparison with the state-of-the-art methods on four benchmarks. $C^3L^*$ denotes the LVLM is fine-tuned using VLIT data generated by the other model ( \textit{e.g.,} MiniGPT-4 w/ $C^3L^*$ denotes that data is generated by LLaVA).}
    \label{tab:table_3}
\end{table*}

\begin{table}[htbp]
\centering
\resizebox{\linewidth}{!}{\begin{tabular}{l*{6}{c}}
\toprule
Dataset & Instances &  Avg. Q len & Avg. A len   \\
\midrule

ComVint~\cite{comvint} & 32k & 23.3 & 18.9  \\  

 LLaVA-158k~\cite{llava} & 158k & 11.0 & 66.7  \\
LRV~\cite{LRV} & 400k & 12.1 & 15.2    \\
\midrule
Ours & 5k  & 23.4 & 12.8  \\
\bottomrule
\end{tabular}}
    \caption{Statistic of VLIT Dataset. “Instances” represents the total number of data instances in the dataset. “Avg. Q len” denotes the average length of questions in the dataset, while “Avg. A len” denotes the average length of answers.}
    \label{tab:table_1}
\end{table}

\paragraph{Contrastive learning module.}
 Due to the presence of “exposure bias” problem, we introduce the contrastive learning module for models to learn experience from low-quality samples in the initial dataset.  
 Specifically, given an image $I_i$, the data sample with the highest $S(I^2C)$ is employed as the positive pseudo-label and the remaining data samples as negative pseudo-labels. 
 By incorporating the positive-negative pseudo-labels and  the contrastive learning framework, our module effectively enhances the capability of the LVLM to generate VLIT data.

\paragraph{Final data generation.} 
  The final 5k VLIT data is generated using the enhanced LVLM along with the two-step generation and can be employed to fine-tune other LVLMs, thereby enhancing the overall capacity of the models.

\section{Experiments}

\subsection{Experimental Setting}

\begin{table}[htbp]
    \centering

\resizebox{\linewidth}{!}{\begin{tabular}{l*{3}{c}}
\toprule
Dataset & Model & SEED  \\
\midrule
ComVint~\cite{comvint} &  LLaVA-7B & 54.21  \\  
 LLaVA-158k~\cite{llava} &  LLaVA-7B & 49.53  \\
LRV~\cite{LRV} &  LLaVA-7B & 56.58   \\
\midrule
Ours &  LLaVA-7B &  56.09 \\
\bottomrule
\end{tabular}}
    \caption{The evaluation results on SEED after fine-tuning LLaVA-7B using different VLIT datasets.}
    \label{tab:table_2}
\end{table}

\paragraph{Implementation details.}
We select two representative LVLMs (\textit{i.e.,} LLaVA~\cite{llava-v1-5} and MiniGPT-4~\cite{minigpt4}) as our backbones to demonstrate the generality of our  method.
During the initial data generation phase, we utilize images from the COCO 2014~\cite{coco-2014} training set and generate 20k initial VLIT data samples.
When fine-tuning backbones using our final data, following ~\cite{llava}, we fix the weights of the CLIP~\cite{clip} visual encoder during fine-tuning. The AdamW~\cite{adam} is used as our optimizer with the weight decay 1e-5. We fine-tune all models   with a learning rate of 2e-5 and batch size is set to 2. For MiniGPT-4, two 3090Ti GPUs are used while for LLaVA we use eight 2080Ti GPUs.

\paragraph{Benchmarks.} 
To comprehensively evaluate the performance of LVLMs after VLIT using data generated by our $C^3L$, we select four benchmarks  (\textit{i.e.,} SEED, MMB, LLaVA$\rm^W$ and POPE) for evaluation.

\noindent 1) SEED~\cite{seed-bench} consists of 19k multiple-choice questions to evaluate the image and video understanding capabilities of LVLMs across 12 different tasks. Following~\cite{complex}, we only evaluate the capacity of LVLMs on image-related tasks.

\noindent 2) MMB~\cite{mmbench} develops a comprehensive evaluation pipeline which incorporates a novel \textit{CircularEval} strategy  to effectively evaluate the capabilities of LVLMs systematically. We use \textit{dev} split of MMB for evaluation.

\noindent 3) LLaVA$\rm^W$~\cite{llava-v1-5} encompasses indoor and outdoor scenes, memes, paintings, sketches, etc, to evaluate the capabilities of LVLMs in more challenging tasks.

\noindent 4) POPE~\cite{pope} aims to effectively evaluate the common issue of hallucinations in almost all LVLMs. By constructing three types of  sampling strategies along with a polling-based query method,  they can accurately measure the severity of model hallucinations for different objects.

\paragraph{Generated dataset.}  
As shown in Table \ref{tab:table_1}, we conduct a statistical analysis of our final VLIT dataset. It is evident that our dataset of 5k instances is  smaller in number compared to other VLIT datasets, such as LRV-400k~\cite{LRV}. In Table \ref{tab:table_2}, it can be observed that  the performance of LLaVA-7B~\cite{llava} when fine-tuned using our 5k data is competitive to the performance achieved when using LRV-400k, suggesting the effectiveness of our $C^3L$.

\subsection{Quantitative Evaluation}
To assess the improvements in the overall capabilities of LVLMs  fine-tuned using data generated by our $C^3L$,  we compare  these fine-tuned LVLMs with the state-of-the-arts models on four benchmarks in Table \ref{tab:table_3}. 
 The $C^3L^*$ denotes that  VLIT data used to fine-tune the LVLM is generated by the other LVLM (\textit{e.g.,} MiniGPT-4 w/ $C^3L^*$ denotes that data is generated by LLaVA). 
 We observe that  LVLMs with $C^3L$ achieve comparable or even better results on these four widely used benchmarks compared to other state-of-the-art models.
 More importantly, only 5k VLIT data generated by models is utilized to fine-tune them.
 On the SEED, LLaVA w/ $C^3L$ outperforms the previous models with an improvement of 2.7$\%$. 
 This demonstrates the effectiveness of our $C^3L$ in improving the content relevance between VLIT data and images, thereby further enhancing the image understanding capabilities of LVLMs. 
 LLaVA$\rm^W$ is designed to evaluate the capacity of LVLMs in more challenging tasks,  we observe that MiniGPT-4 w/ $C^3L$ achieves better performance than previous models on this benchmark.
\begin{table}
    \centering
\resizebox{\linewidth}{!}{\begin{tabular}{c|cccc|ccc}
\toprule
 & instructions &  filtering & CRM  & CLM  & SEED  & POPE \\
\midrule
1   & \checkmark & \checkmark  & -  & - & 53.2  & 73.2 \\
2  & \checkmark & \checkmark  & - & \checkmark & 54.1  & 74.4 \\
3  & \checkmark & \checkmark  & \checkmark &  - & 55.8  & 75.1 \\
4  & \checkmark & \checkmark   & \checkmark &  \checkmark &  \textbf{56.1}  & \textbf{76.0} \\
\bottomrule
\end{tabular}}
    \caption{\textbf{Ablation study.} We ablate key components to demonstrate the effectiveness of our method. The “instructions”  denotes  the instructions used to guide LVLMs to generate data. The “filtering”  represents the basic filtering process that removes duplicate or invalid data identified by heuristics (\textit{e.g.,} data is too short or too long). “CRM” and ``CLM'' are Content Relevance Module and Contrastive Learning Module respectively.}
    \label{tab:table_4}
\end{table}
 On the POPE, models with $C^3L$ achieve comparable results with previous models  which  suggests that data generated by our $C^3L$ exhibits higher content relevance with images, thus can be applied to improve the factualness in LVLMs.

\begin{table}
    \centering
\resizebox{\linewidth}{!}{\begin{tabular}{l*{2}{c}}
\toprule
Method & SEED \\
\midrule
LLaVA-7B w/ Self-Instruct~\cite{self-instruct} & 53.9 \\
LLaVA-7B w/ VIGC~\cite{vigc}  & 55.2 \\
LLaVA-7B w/ $I^2C$ (Ours) & \textbf{56.1} \\
\midrule
LLaVA-13B w/ Self-Instruct~\cite{self-instruct} & 56.6 \\
LLaVA-13B w/ VIGC~\cite{vigc}  & 57.4 \\
LLaVA-13B w/ $I^2C$ (Ours) & \textbf{58.3} \\
\bottomrule
\end{tabular}}
    \caption{\textbf{Alternative data selection methods testing.} We compare with two other data selection methods on SEED to test the effectiveness of our $I^2C$ scores.}
    \label{tab:table_5}
\end{table}
\begin{figure}[h]

\centering

\includegraphics[width = 8.5cm ]{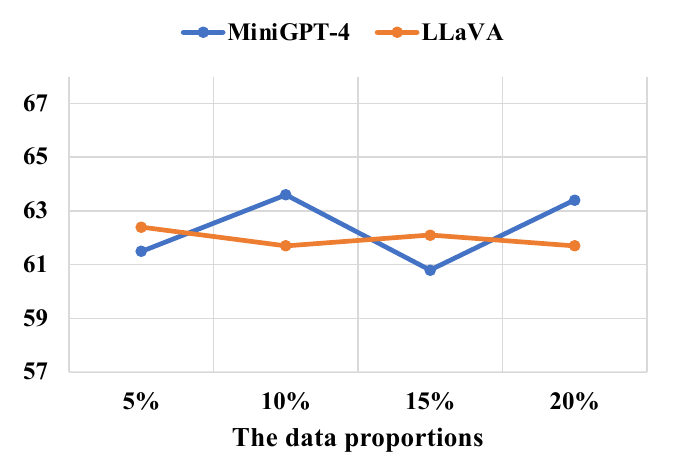}
\caption{\textbf{Alternative data selection proportions testing.} We conduct experiments on LLaVA$\rm ^W$ to test the effects of different data selection proportions.}
\label{fig:fig_3}
\end{figure}

\subsection{Ablation Studies}
  In order to demonstrate the effectiveness of our method in improving the overall capacity of the LVLM including factualness, we conduct several ablation studies on the SEED  and POPE.  As shown in Table \ref{tab:table_4}, 
in the first row, the baseline is established by fine-tuning LLaVA-7B~\cite{llava} using VLIT data generated with instructions and basic filtering.
In the second row, it can be found that the performance on SEED and POPE exhibits an improvement of  0.9$\%$ and 1.2$\%$ respectively.
This could be attributed to the effectiveness of our contrastive learning module in solving the  “exposure bias” problem and enhancing the capacity of the model as a capable data generator. In the third row, we can find that our content relevance module plays a pivotal role, bringing a 2.6$\%$ improvement on SEED  and a 1.9$\%$ improvement on POPE in comparison to the baseline model. These results prove that our content relevance module can enhance the correspondence between VLIT data and images, thus improving the overall capabilities of models when fine-tuned using the data. In the last row, it can be observed that additional improvements are achieved when the two modules work collaboratively, which demonstrate that our $C^3L$ is remarkably effective in improving the content relevance of the model-generated VLIT data with images and enhancing the capability of the model as a data generator.

\begin{figure*}[h]

\centering

\includegraphics[width = 17.5cm]{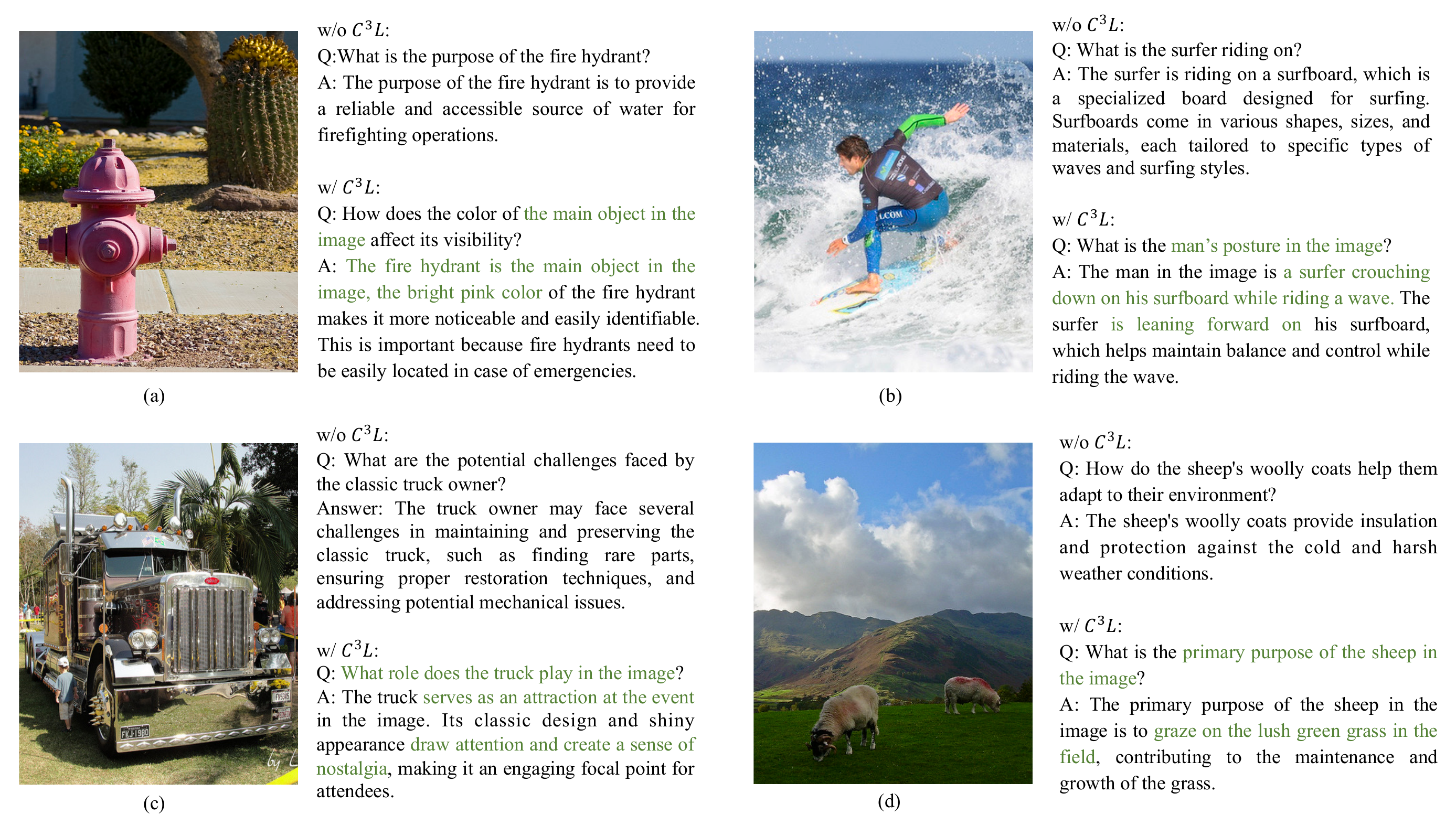}
\caption{\textbf{Generation results.} We show the VLIT data generated w/o $C^3L$ and w/ $C^3L$.}
\label{fig: fig_4}
\end{figure*}

\subsection{Alternative Data Selection Setting}
We propose two alternative data selection settings and conduct several experiments to measure the  reliability of our method.  

\paragraph{Alternative  data selection  methods.} To further demonstrate the effectiveness of our $S(I^2C)$ in selecting positive/negative pseudo-labels, we compare our method with two other approaches (\textit{i.e.,} Self-Instruct~\cite{self-instruct} and VIGC~\cite{vigc}) on SEED. As shown in Table \ref{tab:table_5}, we conduct experiments by replacing the I2C scores with these two methods using two backbones (\textit{i.e.,} LLaVA-7B and LLaVA-13B). For Self-Instruct, we apply the same filtering and post-processing methods following~\cite{self-instruct} to identify low-quality data, the remaining instruction samples are considered as positive pseudo-labels. For VIGC, we generate data using two approaches: direct generation and iterative generation~\cite{vigc}. The data generated by iterative generation is considered as positive pseudo-labels.
The results show that selecting positive/negative pseudo-labels by $S(I^2C)$ is more effective compared to other alternative data selection methods.

\paragraph{Alternative data selection proportions.}
In order to investigate the impact of the 
data proportions selected by content relevance module on the performance of the LVLMs (\textit{i.e.,} LLaVA and MiniGPT-4) fine-tuned with $C^3L$, we compare the performance of LVLMs on LLaVA$\rm^W$ after fine-tuning with different data proportions. As shown in Fig. \ref{fig:fig_3}, the blue line denotes the performance  of MiniGPT-4  changes with different data selection proportions, while the  orange line corresponds to LLaVA. The results demonstrate that there is a marginal difference in the performance of LVLMs when fine-tuned with different data proportions. Therefore, we choose  10$\%$ data in the content relevance module in order to balance performance and cost.

\subsection{Qualitative Evaluation}

As shown in Fig. \ref{fig: fig_4}, we show some qualitative results generated by $C^3L$. 
Through overall comparison, the data generated by our $C^3L$ exhibits higher content relevance with images. For example, in Fig. \ref{fig: fig_4} (a),
answering the question generated without $C^3L$
 does not require leveraging information from the image, as the fire hydrant is primarily used to supply water during fire incidents. On the contrary, answering the question generated by $C^3L$ not only requires common knowledge about the fire hydrant but also necessitates a comprehensive understanding of the image. Additionally, the data generated by $C^3L$ is also more detailed and factual. In Fig. \ref{fig: fig_4} (b)-(d), the generated data containing the question about the man's posture and the answer about the role of the truck, which is “serves as an attraction at the event”.

\section{Conclusion}
In this paper, we propose a new  Content Correlated Vision-Language Instruction Tuning Data Generation via
Contrastive Learning ($C^3L$). We first design a new content relevance module to  improve the correlation between VLIT data and images. Meanwhile, we propose a contrastive learning module,  which utilizes the generated samples as pseudo-labels to boost the data generation capacity of the LVLMs.
  Furthermore,  an augmented VLIT data generation pipeline is proposed by combining these two modules, which can be applied to generate VLIT data that exhibits high content relevance with images.
  According to automatic evaluations,  fine-tuned LVLMs using data generated by our $C^3L$ achieve comparable  performance to the state-of-the-art models on four benchmarks and
provides a new paradigm for our community.

\section*{Contribution Statement}
Ji Ma and Wei Suo contribute equally to this work. Peng Wang is the corresponding author.

\newpage

\section*{Acknowledgements}
This work was supported by the National Natural Science Foundation of China (No.U23B2013), Shaanxi Provincial Key R\&D Program (No.2021KWZ-03), and Natural Science Basic Research Program of Shaanxi (No.2021JCW-03).

\bibliographystyle{named}
\bibliography{ijcai24}

\end{document}